\documentclass[11pt,letterpaper]{article}
\usepackage[hyperref]{naaclhlt2018}
\usepackage{times}
\usepackage{latexsym}
\usepackage{multirow}
\usepackage{graphicx}
\usepackage{color}
\usepackage{makecell}

\definecolor{darkgreen}{rgb}{0.0, 0.5, 0.0}

\aclfinalcopy 
\def\aclpaperid{***} 

\title{Multi-lingual neural title generation for e-Commerce browse pages}

\author{
    Prashant Mathur \and Nicola Ueffing \and Gregor Leusch \\
    MT Science Team \\
  	eBay\\
  	Kasernenstra{\ss}e 25\\
  	Aachen, Germany}

\date{\today}
\begin{document}

\maketitle

\begin{abstract}
To provide better access of the inventory to buyers and better search engine optimization, e-Commerce websites are automatically generating millions of easily searchable browse pages.
A browse page consists of a set of slot name/value pairs within a given category, grouping multiple items which share some characteristics.
These browse pages require a title describing the content of the page. 
Since the number of browse pages are huge, manual creation of these titles is infeasible.
Previous statistical and neural approaches depend heavily on the availability of large amounts of data in a language.
In this research, we apply sequence-to-sequence models to generate titles for high- \& low-resourced languages by leveraging transfer learning.
We train these models on multi-lingual data, thereby creating one joint model which can generate titles in various different languages.
Performance of the title generation system is evaluated on three different languages; English, German, and French, with a particular focus on low-resourced French language.

\end{abstract}

\section{Introduction} \label{sec:intro}
Natural language generation (NLG) has a broad range of applications, from question answering systems to story generation, summarization etc. 
In this paper, we target a particular use case that is important for e-Commerce websites, which group multiple items on common pages called \textit{browse pages} (BP).
Each browse page contains an overview of various items which share some characteristics expressed as slot/value pairs.

For example, we can have a browse page for Halloween decoration, which will display different types like lights, figurines, and candy bowls. 
These different items of decoration have their own browse pages, which are linked from the BP for Halloween decoration. 
A ceramic candy bowl for Halloween can appear on various browse pages, e.g.~on the BP for Halloween decoration, BP for Halloween candy bowls, as well as the (non Halloween-specific) BP for ceramic candy bowls.

To show customers which items are grouped on a browse page, we need a human-readable title of the content of that particular page.
Different combinations of characteristics bijectively correspond to different browse pages, and consequently to different browse page titles.

Note that here, different from other natural language generation tasks described in the literature,
slot names are already given; the task is to generate a title for a set of slots.
Moreover, we do not perform any selection of the slots that the title should realize; but all slots need to be realized in order to have a unique title.
E-Commerce sites may have tens of millions of such browse pages in many different languages. 
The number of unique slot-value pairs are in the order of hundreds of thousands.
All these factors render the task of human creation of BP titles infeasible.

\citet*{Mathur:2017} 
developed several different systems which generated titles for these pages automatically. 
These systems include rule-based approaches, statistical models, and combinations of the two. 
In this work, we investigate the use of neural sequence-to-sequence models for browse page title generation. 
These models have recently received much attention in the research community, and are becoming the new state of the art in machine translation (refer Section~\ref{sec:seq2seq}).

We will compare our neural generation models against two state-of-the-art systems.
\begin{enumerate}
\item The baseline system for English and French implements a {\it hybrid} generation approach, which combines a rule-based approach (with a manually created grammar) and statistical machine translation (SMT) techniques. For French, we have monolingual data for training language model, which can be used in the SMT system. 
For English, we also have human-curated titles and can use those for training additional ``translation'' components for this hybrid system.
\item The system for German is an {\it Automatic Post-Editing (APE)} system -- first introduced by~\cite{Simard:07} -- which generates titles with the rule-based approach, and then uses statistical machine translation techniques for automatically correcting the errors made by the rule-based approach.
\end{enumerate}


\section{Related work} \label{sec:relwrk}

The first works on NLG were mostly focused on rule-based language generation~\citep{dale1998realities,Reiter:05,Green:2006}.
NLG systems typically perform three different steps: \textit{content selection}, where a subset of relevant slot/value pairs are selected, followed by \textit{sentence planning}, where these selected pairs are realized into their respective linguistic variations, and finally \textit{surface realization}, where these linguistic structures are combined to generate text.
Our use case differs from the above in that there is no selection done on the slot/value pairs, but all of them undergo the sentence planning step.
In rule-based systems, all of the above steps rely on hand-crafted rules. 


The recent work focuses on generating texts from structured data as input by performing selective generation, i.e.~they run a selection step that determines the slot/value pairs which will be included in the realization~\cite{Mei:15,Lebret:16,duma:13,chisholm:17}. 
In our use case, all slot/value pairs are relevant and need to be realized.

\citet{serban2016generating} generate questions from facts (structured input) by leveraging fact embeddings and then employing placeholders for handling rare words.
In their work, the placeholders are heuristically mapped to the facts, however, we map our placeholders depending on the neural attention (for details, see Section~\ref{sec:seq2seq}).


\section{Lexicalization} \label{sec:lexicalization}
Our first step towards title generation is verbalization of all slot/value pairs.
This can be achieved by a rule-based approach as described in 
{\it anonymous citation}.
However, in the work presented here, we do not directly lexicalize the slot/value pairs, but realize them in a pseudo language first. 
For example, the pseudo-language sequence for the slot/value pairs in Table~\ref{tab:bn-example} is
``{\it \_brand} ACME {\it \_cat} Cell Phones \& Smart Phones {\it \_color} white {\it \_capacity} 32GB''.\footnote{\_cat refers to the selected category in the browse page.}



\begin{table}
\footnotesize
    \begin{centering}
    \begin{tabular}{|l|l|}
    \hline
    Slot Name & Value \\
    \hline
    Category & \textit{Cell Phones \& Smart Phones}\\
    Brand& \textit{ACME}\\
    Color& \textit{white}\\
    Storage Capacity& \textit{32GB}\\
    \hline
    \end{tabular}
    \caption{\footnotesize{Example of browse page slot/value pairs.}}
    \label{tab:bn-example}
    \end{centering}
\end{table}

\subsection{Normalization} \label{subsec:normalization}
Pseudo-language browse pages can still contain a large number of unique slot values. For example, there exist many different brands for smart phones (Samsung, Apple, Huawei, etc.).
Large vocabulary is a known problem for neural systems, because rare or less frequent words tend to translate incorrectly due to data sparseness~\cite{LuongRare15}.
At the same time, the softmax computation over the large vocabulary becomes intractable in current hardware.
To avoid this issue, we normalize the pseudo-language sequences and thereby reduce the vocabulary size.
For each language, we computed 
the 30 most frequent slot types
and normalized their values via placeholders~\cite{LuongPM15}.
For example, a lexicalization of ``\textit{Brand: ACME}'' is ``$\_brand$ ACME'', but after normalization, this becomes $\_brand~\$brand|$\textit{ACME}.
This representation means that the slot type $\_brand$ has the value of a placeholder $\$brand$ which contains the entity called ``ACME''.
During training, we remove the entity from the normalized sequence, while keeping them during translation of development or evaluation set.
The mapping of placeholders in the target text back to entity names is described in Section \ref{sec:seq2seq}.

The largest reduction in vocabulary size would be achieved by normalizing all slots. 
However, this would create several issues in generation.
Consider the pseudo-language sequence ``$\_bike$ Road bike $\_type$ Racing''. If we replace all slot values with placeholders, i.e.~``$\_bike$ \$bike $\_type$ \$type'', then the system will not have enough information for generating the title ``Road racing bike''. Moreover, the boolean slots, such as ``$\_comic$ Marvel comics $\_signed$ No'' would be normalized to placeholders as ``$\_comic$ \$comic $\_signed$ \$signed'', and we would loose the information (``No'') necessary to realize this title as ``Unsigned Marvel comics''.

\subsection{Sub-word units}

We applied another way of reducing the vocabulary, called byte pair encoding (BPE)~\cite{Sennrich15}, a technique often used in NMT systems~\cite{WMT:2017}.
BPE is essentially a data compression technique which splits each word into sub-word units and allows the NMT system to train on a smaller vocabulary.
One of the advantages of BPE is that it propagates generation of unseen words (even with different morphological variations).
However, in our use case, this can create issues, because if BPE splits a brand and generates an incorrect  brand name in the target, an e-Commerce company could be legally liable for the mistake.
In such case, one can first run the normalization with placeholders tags followed by BPE, but due to time constraints, we do not report experiments on the same.


\section{Sequence-to-Sequence Models} \label{sec:seq2seq}

Sequence-to-sequence models in this work are based on an encoder-decoder model and an attention mechanism as described by~\citet{bahdanau2014neural}.
In this network, the encoder is a bi-directional RNN which encodes the information of a sentence $X = (x_1, x_2, \ldots x_m)$ of length $m$ into a fixed length vector of size $|h_i|$, where $h_i$ is the hidden state produced by the encoder for token $x_i$.
Since our encoder is a bi-directional model, the encoded hidden state is $h_i = h_{i,fwd} + h_{i,bwd}$, where $h_{fwd}$ and $h_{bwd}$ are unidirectional encoders, running from left to right and right to left, respectively. 
That is, they are encoding the context to the left and to the right of the current token.

Our decoder is a simple recurrent neural network (RNN) consisting of gated recurrent units (GRU)~\citep{ChoMBB14} because of their computationally efficiency. 
The RNN predicts the target sequence $Y = (y_1, y_2, \ldots y_j \ldots y_l)$ based on the final encoded state $h$.
Basically, the RNN predicts the target token $y_j \in V$ (with target vocabulary V) and emits a hidden state $s_j$ based on the previous recurrent state $s_{j-1}$, the previous sequence of words $Y_{j-1} = (y_1, y_2, \ldots y_{j-1})$ and $C_j$, a weighted attention vector.
The attention vector is a weighted average of all the hidden source states $h_i$, where $i=1,\dots,m$.
Attention weight ($a_{ij}$) is computed between the hidden states $h_i$ and $s_j$ and is leveraged as a weight of that source state $h_i$.
In generation, we make use of these alignment scores to align our placeholders.\footnote{These placeholders are not to be confused with the placeholder for a tensor.}
The target placeholders are bijectively mapped to those source placeholders whose alignment score ($a_{ij}$) is the highest at the time of generation.

The decoder predicts a score for all the tokens in the target vocabulary, which is then normalized by a softmax function, and the token with the highest probability is predicted.

\section{Multilingual Generation} \label{sec:mutlinlg}

In this section, we present the extension of our work from a single-language setting to multi-language settings.
There have been various studies in the past that target neural machine translation from multiple source languages to single target language \citep{ZophK16}, from single source to multiple target languages \citep{DongWHYW15} and multiple source to multiple target languages \citep{johnson2016google}.
One of the main motivation of joint learning in above works is to improve the translation quality on a low-resource language pair via transfer learning between related languages.
For example, \citep{johnson2016google} had no parallel data available to train a Japanese-to-Korean MT system, but training Japanese-English and English-Korean language pairs allowed their model to learn translations from Japanese to Korean without seeing any parallel data. 
In our case, the amount of training data for French is small compared to English and German (cf.~Section~\ref{subsec:data}).
We propose joint learning of English, French and German, because we expect that transfer learning will improve generation for French.
We investigate the joint training of pairs of these languages as well the combination of all three.

On top of the multi-lingual approach, we follow the work of \citet{Currey:2017} who proposed copying monolingual data on both side (source and target) as a way to improve the performance of NMT systems on low-resource languages.
In machine translation, there are often named entities and nouns which need to be translated verbatim, and this copying mechanism helps in identifying them.
Since our use case is monolingual generation, we expect a large gain from this copying approach because we have many brands and other slot values which needs to occur verbatim in the generated titles.

\section{Experiments} \label{sec:exp}
\subsection{Data} \label{subsec:data}

We have access to a large number of human-created titles (curated titles) for English and German and a small number of curated titles for French.
When generating these titles, human annotators were specifically asked to realize all slots in the title.

We make use of a large monolingual out-of-domain corpus for French, as it is a low-resource language.
We collect item description data from an e-Commerce website and clean the data in the following way: 1) we train a language model (LM) on the small amount of French curated titles, 2) we tokenize the out-of-domain data, 3) we remove all sentences with length less than 5, 4) we compute the LM perplexity for each sentence in the out-of-domain data, 5) we sort the sentences in increasing order of their perplexities and 6) select the top 500K sentences.
Statistics of the data sets are reported in Table~\ref{table:stats}.

\begin{table}[htbp]
\small
\begin{center}
\begin{tabular}{l|c|c|c}
\hline
\bf Languages & \bf Set & \bf \#Titles & \bf \#trg Tokens \\
\hline
\multirow{3}{*}{English} & Train & 222k & 1.5M \\ 
& Dev & 1000 & 6682 \\
& Test & 1000 & 6633 \\ \hline
\multirow{3}{*}{German} & Train & 226k & 1.9M \\
& Dev & 1000 & 8876 \\
& Test & 500 & 4414 \\ \hline
\multirow{3}{*}{French} & Train & 10k & 95k \\ 
& Monolingual & 500k & 5.54M \\
& Dev & 486 & 6403 \\
& Test & 478 & 3886 \\
\hline
\end{tabular}
\end{center}
\caption{\label{table:stats} \footnotesize{Training data statistics per language.
`k' and `M' stands for thousand and million, respectively.}}
\end{table}

\subsection{Systems} \label{subsec:systems}
We compared the NLG systems in the single-, dual-, and multi-lingual settings.

\paragraph{Single-language setting:} This is the baseline NLG system, a straightforward sequence-to-sequence model with attention as described in~\citet{LuongPM15}, trained separately for each language. %
The vocabulary is computed on the concatenation of both source and target data, and the same vocabulary is used for both source and target languages in the experiments.

We use Adam~\citep{KingmaB14} as a gradient descent approach for faster convergence.
Initial learning rate is set to 0.0002 with a decay rate of 0.9.
The dimension of word embeddings is set to 620 and hidden layer size to 1000.
Dropout is set to 0.2 and is activated for all layers except the initial word embedding layer, because we want to realize all aspects, we cannot afford to zero out any token in the source.
We continue training of the model and evaluate on the development set after each epoch, stopping the training if the BLEU score on the development set does not increase for 10 iterations.

\paragraph{Baselines:} 
We compare our neural system with a \textbf{fair} baseline system (\textit{Baseline 1}), which is a statistical MT system trained on the same parallel data as the neural system: the source side is the linearized pseudo-language sequence, and the target side is the curated title in natural language. 
\textit{Baseline 2} is the either the hybrid system (for French and English) or the APE system (for German), both described in Section~\ref{sec:intro}. These are \textbf{unfair} baselines, because (1) the hybrid system employs a large number of hand-made rules in combination with statistical models
~\citep*{Mathur:2017},
while the neural systems are unaware of the knowledge encoded in those rules, (2)
the APE system and
neural systems are learn from same \textit{amount} of parallel data, but the APE system aims at correcting rule-based generated titles, whereas the neural system aims at generating titles directly from a linearized form, which is a harder task. 
As in the paper, we compare with the best performing systems i.e. hybrid system for English and French, and APE system for German.

\paragraph{Multi-lingual setting:}
We train the neural model jointly on multiple languages to leverage transfer learning from a high-resource language to a low-resource one.
In our multi-lingual setting, we experiment with three different combinations to improve models for French: 1) English+French (\emph{en-fr}) 2) German+French (\emph{de-fr}) 3) English+French+German (\emph{en-fr-de}). 
English and French being close languages, we expect the \emph{en-fr} system to benefit more from transfer learning across languages than any other combination.
Although, as evident in~\citet{ZophK16}, joint learning between the distant languages works better as they tend to disambiguate each other better than two languages which are close.
For comparison, we also run a combination of two high-resource languages, i.e.~English and German (\emph{en-de}), to see if transfer learning works for them.
It is important to note that in all multi-lingual systems the low-resourced language is over-sampled to balance the data.

We used the same design parameters on the neural network in both the single-language and the multi-lingual setting. 

\paragraph{Normalized setting:}
On top of the systems above, we also experimented with the normalization scheme presented in Section~\ref{subsec:normalization}.
Normalization is useful in two ways: 1) It reduces the vocabulary size and 2) it avoids spurious generation of important aspect values (slot values).
The second point is especially important in our case because this avoids highly sensitive issues such as brand violations.
MT researches have observed that NMT systems often generate very fluent output, but have a tendency to generate inadequate output, i.e.~sentences or words which are not related to the given input~\citep{Koehn:17}.
We alleviate this problem through the normalization described above. After normalization we see vocabulary reductions of 15\% for French, 20\% for German and as high as 35\% for English.

As described in Section~\ref{sec:mutlinlg}, we also use byte pair encoding, with a BPE code size of 30,000 for all systems (with BPE). 
We train the codes on the concatenation of source and target since (being monolingual) the vocabularies are very similar; the vocabulary size is around 30k for systems using BPE for both source and target.

\section{Results} \label{sec:res}

We evaluate our systems with three different automatic metrics: BLEU~\citep{BLEU-2002-Papineni}, TER~\citep{Snover06astudy} and character F-Score~\citep{popovic:2016:WMT}.
Note that BLEU and character F-score are quality metrics, i.e.~higher scores mean higher quality, while TER is an error metric, where lower scores indicate higher quality.
All metrics compare the automatically generated title against a human-curated title and determine sequence matches on the word or character level.


\begin{table}[htbp]
\small
\begin{center}
\begin{tabular}{llccc}
\hline
\bf System & \bf Norm. & \bf BLEU$\uparrow$ & \bf chrF1$\uparrow$ & \bf TER$\downarrow$ \\
\hline
Baseline 1 & n/a & 64.2 & 82.9 & 26.5 \\
Baseline 2 & n/a & 74.3 & 86.1 & 19.8 \\
\hline
$en$ & No & 68.4 & 82.8 & 21.2 \\
$en$ & Yes(Tags) & 67.1 & 82.5 & 21.7 \\
\hline
$en$-$fr$ & No & 70.7 & 83.9 & 20.1 \\
$en$-$fr$ & Yes(Tags) & 67.1 & 82.1 & 22.8 \\
$en$-$fr$ & Yes(BPE) & 71.9 & 85.2 & 18.5 \\
$en$-$fr_{big}$ & Yes(BPE) & 74.1 & 86.2 & 17.3 \\ \hline
$en$-$de$ & No & 65.8 & 80.7 & 23.6 \\
$en$-$de$ & Yes(Tags) & 67.1 & 82.8 & 22.3 \\
$en$-$de$ & Yes(BPE) & 72.7 & 85.4 & 18.8 \\
\hline
$en$-$fr$-$de$ & Yes(BPE) & 74.5 & 86.3 & 17.0 \\
\hline
\end{tabular}
\end{center}
\caption{\label{table:res:english}\footnotesize{Results on EN test, cased and detokenized.}}
\end{table}

Table~\ref{table:res:english} summarizes results from all systems on the {\bf English} test set. 
All neural systems are better than the fair \textit{Baseline 1} system.

Normalization with tags (i.e.~using placeholders) has a negative effect on English title quality both in the single-language setting $en$ (67.1 vs.~68.4 BLEU) and in the dual-language setting $en$-$fr$ (67.1 vs.~70.7 BLEU). 
However, title quality increases when using BPE instead (71.9 vs.~70.7 BLEU).
On $en$-$de$, we observe gains both from normalization with tags and from BPE.
Again, BPE normalization works best.
Both dual-language systems with BPE achieve better performance that the best monolingual English system (71.9 and 72.7 vs.~68.4 BLEU).

The system $en$-$fr_{big}$ 
contains monolingual French data added via the copying mechanism, which improves title quality.
It outperforms any other neural system and is on par with \textit{Baseline 2} (unfair baseline), even outperforming it in terms of TER.
The multi-lingual system $en$-$fr$-$de$ is very close to $en$-$fr_{big}$ according to all three metrics.



\begin{table}[htbp]
\small
\begin{center}
\begin{tabular}{llccc}
\hline
\bf System & \bf Norm. & \bf BLEU$\uparrow$ & \bf chrF1$\uparrow$ & \bf TER$\downarrow$ \\
\hline
Baseline 1 & n/a & 58.5 & 88.3 & 31.4 \\
Baseline 2 & n/a & 79.4 & 90.7 & 17.1 \\
\hline
$de$ & No & 78.2 & 87.0 & 20.7 \\
$de$ & Yes(Tags) & 71.1 & 85.0 & 27.2 \\
\hline
$en$-$de$ & No & 74.0 & 87.3 & 22.6 \\
$en$-$de$ & Yes(Tags) & 65.6 & 84.0 & 30.2 \\
$en$-$de$ & Yes(BPE) & 79.6 & 91.1 & 16.6 \\ \hline
$de$-$fr$ & No & 77.2 & 88.9 & 18.9 \\
$de$-$fr$ & Yes(Tags) & 63.3 & 83.0 & 30.7 \\
$de$-$fr$ & Yes(BPE) & 77.6 & 89.0 & 19.2 \\
$de$-$fr_{big}$ & Yes(BPE) & 80.0 & 91.6 & 16.2 \\
\hline
$en$-$fr$-$de$ & Yes(BPE) & 80.6 & 92.0 & 15.3 \\
\hline
\end{tabular}
\end{center}
\caption{\label{table:res:german}\footnotesize{Results on DE test, cased and detokenized.}}
\end{table}

Table~\ref{table:res:german} collects the results for all systems on the {\bf German} test set.
For the single-language setting, we see a loss of 7 BLEU points when normalizing the input sequence, which is caused by incorrect morphology in the titles. 
When using placeholders, 
the system generates entities in the title in the exact form in which they occur in the input. 
In German, however, the words often need to be inflected.
For example, the slot ``{\it \_brand} Markenlos'' should be realized as ``Markenlose'' (Unbranded) in the title, but the placeholder 
generates the input form ``Markenlos'' (without suffix 'e'). 
This causes a huge deterioration in the word-level metrics BLEU and TER, but not as drastic in chrF1, which evaluates on the character level.

For German, there is a positive effect of transfer learning for both dual-language systems $en$-$de$ and $de$-$fr_{big}$ with BPE  (79.6 and 80.0 vs.~78.2 BLEU).
However, the combination of languages hurts when we combine languages at token level, i.e.~without normalization or with tags. 
The performance of systems with BPE is even on par with the strong baseline of 79.4 BLEU, both for combinations of two and of three languages.



\begin{table}[htbp]
\small
\begin{center}
\begin{tabular}{llccc}
\hline
\bf System & \bf Norm. & \bf BLEU$\uparrow$ & \bf chrF1$\uparrow$ & \bf TER$\downarrow$ \\
\hline
Baseline 1 & n/a & 44.6 & 77.7 & 44.3 \\
Baseline 2 & n/a & 76.8 & 89.0 & 18.4 \\
\hline
$fr_{small}$ & No & 23.0 & 52.0 & 71.1 \\
$fr_{small}$ & Yes(Tags) & 27.4 & 56.2 & 60.1 \\
$fr_{big}$ & Yes(BPE) & 29.5 & 57.3 & 58.5 \\
$fr_{big}$ & Yes(Both) & 31.4 & 61.3 & 60.9 \\
\hline
$en$-$fr$ & No & 22.5 & 51.3 & 69.6 \\
$en$-$fr$ & Yes(Tags) & 20.1 & 47.1 & 70.3 \\
$en$-$fr$ & Yes(BPE) & 21.6 & 50.7 & 73.9 \\
$en$-$fr_{big}$ & Yes(BPE) & 32.6 & 61.8 & 51.2 \\ \hline
$de$-$fr$ & No & 21.7 & 50.2 & 71.4 \\
$de$-$fr$ & Yes(Tags) & 23.2 & 49.9 & 67.3 \\
$de$-$fr$ & Yes(BPE) & 30.9 & 63.0 & 61.8 \\
$de$-$fr_{big}$ & Yes(BPE) & 38.8 & 67.8 & 50.5 \\
\hline
$en$-$fr$-$de$ & Yes(BPE) & 45.3 & 73.2 & 42.0 \\
\hline
\end{tabular}
\end{center}
\caption{\label{table:res:french}\footnotesize{Results on FR test, cased and detokenized.}}
\end{table}

Table~\ref{table:res:french} summarizes the results from all systems on the {\bf French} test set.
The single-language $fr$ NMT system achieves a low BLEU score compared to the SMT system Baseline 1 (23.0 vs.~44.6).
This is due to the very small amount of parallel data, which is a setting where SMT typically outperforms NMT as evidenced in~\citet{zoph2016transfer}.
Normalization has a big positive impact on all French systems (e.g.~23.0 vs.~27.4 BLEU for $fr$).

The $de$-$fr$ systems show a much larger gain from transfer learning than the 
$en$-$fr$ systems, which validates~\citet{ZophK16}'s results, who show that transfer learning is better for distant languages than for similar languages.

For all three languages, copying monolingual data improves the NMT system by a large margin.

The multi-lingual $en$-$fr$-$de$ (BPE) system (with copied monolingual data) is the best system for all three languages. 
It has the additional advantage of being one single model that can cater to all three languages at once.

\begin{table} [htbp]
\small
    \begin{centering}
    \begin{tabular}{|l|p{0.37\textwidth}|}
    \hline
    System & Title \\
    \hline
    src & \_cat \'Equipements de garage \_brand Outifrance \\
    ref & \textcolor{darkgreen}{\'Equipements de garage Outifrance} \\
    $fr_{small}$ & \textcolor{darkgreen}{\'Equipements de} \textcolor{red}{suspension et de travail} \\
    $fr_{small,tags}$ & \textcolor{darkgreen}{\'Equipements de garage Outifrance} \\ \hline
    src & \_cat Cylindres \'emetteurs d'embrayage pour automobiles \_brand Vauxhall \\
    ref & \textcolor{darkgreen}{Cylindres \'emetteurs d'embrayage pour automobiles Vauxhall} \\
    $fr_{small}$ & \textcolor{red}{Perles} \textcolor{darkgreen}{d'embrayage pour} \textcolor{orange}{automobile} \textcolor{darkgreen}{Vauxhall} \\
    $fr_{big}$ & \textcolor{darkgreen}{Cylindres \'emetteurs d'embrayage pour} \textcolor{orange}{automobile} \textcolor{darkgreen}{Vauxhall} \\ \hline
    src & \_cat Dessous de verre de table \_brand Amadeus \\
    ref & \textcolor{darkgreen}{Dessous de verre de table Amadeus} \\
    $fr_{big}$ & \textcolor{red}{Guirlandes} \textcolor{darkgreen}{de verre} \textcolor{red}{Dunlop} \textcolor{purple}{de table} \\
    $en$-$fr$-$de$ & \textcolor{darkgreen}{Dessous de verre de table Amadeus} \\ \hline
    \end{tabular}
    \caption{\footnotesize{Examples from the french test set.}}
    \label{tab:examples}
    \end{centering}
\end{table}

Table~\ref{tab:examples} present the example titles comparing different phenomenon.
The first block shows the usefulness of placeholders in system $fr_{small,tags}$ (i.e. $fr_{small}$, normalized with tags) where in comparison to $fr_{small}$ the brand  is generated verbatim.
The second block shows the effectiveness of copying the data where ``Cylindres'' is generated correctly in the $fr_{big}$ (with BPE) system in comparison to $fr_{small}$.
Last block shows that reordering and adequacy in generation can be improved with the helpful signals from high resourced English and German languages.


\section{Conclusion} \label{sec:conclusion}
We developed neural language generation systems for an e-Commerce use case for three languages with very different amounts of training data and observed the following:
(1) The lack of resources in French leads to generation of low quality titles, but this can be drastically improved upon with transfer learning between French and English and/or German. 
(2) In case of low-resource languages, copying monolingual data (even if out-of-domain) improves the performance of the system.
(3) Normalization with placeholders usually helps for languages with relatively easy morphology. 
(4) It is important to over-sample the low-resourced languages in order to balance the high- \& low-resourced data, thereby, creating a stable NLG system.
(5) For French, a low-resource language in our use case, the hybrid system which combines manual rules and SMT technology is still far better than the best neural system.
(6) The multi-lingual model has the best trade-off, as it achieves the best results among the neural systems in all three languages and it is one single model which can be deployed easily on a single GPU machine.

\section*{Acknowledgments}
Thanks to Pavel Petrushkov for all the help with the neural MT toolkit.

\bibliography{naaclhlt2016}
\bibliographystyle{acl_natbib}

\end{document}